\documentclass{article} % For LaTeX2e
\usepackage{iclr2022_conference,times}

\usepackage{amsmath,amsfonts,bm}

\def\eqref#1{equation~\ref{#1}}
\def\1{\bm{1}}

\newcommand{\test}{\mathcal{D_{\mathrm{test}}}}

\DeclareMathAlphabet{\mathsfit}{\encodingdefault}{\sfdefault}{m}{sl}
\SetMathAlphabet{\mathsfit}{bold}{\encodingdefault}{\sfdefault}{bx}{n}

\usepackage{hyperref}
\usepackage{url}
\usepackage{amsmath,amssymb,amsfonts}
\usepackage{graphicx}
\usepackage{textcomp}
\usepackage{xcolor} 
\usepackage{epsfig}
\usepackage[font=small,skip=0pt]{caption}
\usepackage{amsmath,amssymb,amsthm}
\usepackage{comment} 
\usepackage{tabularx}
\usepackage{graphicx}
\usepackage{enumitem}
\usepackage{eufrak}
\usepackage{lipsum}
\usepackage{stfloats}
\usepackage{bbm}
\usepackage{relsize}
\usepackage{multicol}
\usepackage[ruled]{algorithm2e}

\newtheorem{thm}{Theorem}[section]
\newtheorem{lem}{Lemma}[section]

\newtheorem{corollary}{Corollary}[section]

\title{A Scale-Independent Multi-Objective Reinforcement Learning with Convergence Analysis}

\author{Mohsen Amidzadeh  \\
	Department of Information and Communications Enginnering\\
	Aalto Universiy, Finland \\
	mohsen.amidzade@aalto.fi 
}
\iclrfinalcopy % Uncomment for camera-ready version, but NOT for submission.
\begin{document}

	\maketitle
	
	\begin{abstract}
		Many sequential decision-making problems need optimization of different objectives which possibly conflict with each other.
		The conventional way to deal with a multi-task problem is to establish a scalar objective function 
		based on a linear combination of different objectives.
		However, for the case of having conflicting objectives with different scales, 
		this method needs a trial-and-error approach to properly find proper weights for the combination. 
		As such, in most cases, this approach cannot guarantee an optimal Pareto solution.
		In this paper, we develop a single-agent scale-independent multi-objective reinforcement learning 
		on the basis of the Advantage Actor-Critic (A2C) algorithm.
		A convergence analysis is then done for the devised multi-objective algorithm providing a convergence-in-mean guarantee.
		We then perform some experiments over a multi-task problem to evaluate the performance of the proposed algorithm.
		Simulation results show the superiority of developed multi-objective A2C approach against the single-objective algorithm.
	\end{abstract}

\section{Introduction}
Many sequential decision-making problems include multiple objective functions competing with each other.
The common approach to finding an optimum solution for these problems
is a scalarization approach based on considering a preference for different objectives. % \citep{?}.
However, the Pareto solutions cannot be  obtained via this method \citep{Kirlik2014}.
As such, a trial-and-error approach might be needed to tune optimum scalarization settings.
This difficulty also appears for the sequential decision-making problems 
being modeled based on a Multi-Objective Markov Decision Process (MO-MDP) \citep{Roijers2013}.
To find an optimum policy for a MO-MDP with different tasks, 
a Multi-Objective Reinforcement Learning (MO-RL) algorithm needs to be developed.
A typical method for MO-RLs is the scalarization approach to first construct a scalar reward based on a combination of
competing rewards (either linear or non-linear) and then apply a single-objective RL algorithm \citep{Moffaert2013,Natarajan2005}.
However, this approach mainly makes the solution highly dependent on the selected combination.

Most developed MO-RL algorithms are restricted to the discrete environment.
\citet{Iima2014} consider a  multi-objective Bellman operator by which a
value-based reinforcement learning algorithm is devised to obtain the Pareto solutions in a discrete environment.
\citet{Mossalam2016} develop a MO-RL on the basis of deep Q-learning and optimistic linear support learning.
They consider a scalarized vector and potential optimal solutions to have a convex combination of the objectives.
However, they need to search over all potential scalarizing vectors as the importance of distinct objectives 
is not a priori knowledge.
\citet{Yang2019} leverage a multi-objective variant of Q-learning with a single-agent approach 
in order to learn a preference-based adjustment being generalized across different preferences.
However, they use a convex envelope of the Pareto frontier during updating process. 
This is often sample inefficient and leads to a sub-optimal policy, though it is efficient from the computational complexity.
Apart from the approach for discrete state-action spaces, 
there exist MO-RL algorithms devised for the continuous environment.
\citet{Chen2019} devise a MO-RL by meta-learning.
They learned a meta-policy distribution trained with multiple tasks.
The multi-objective problem is converted to a number of single-objective problems using a parametric scalarizing function.
However, the solution depends on the distribution based on which the parameters of the scalarizing function are drawn.
Reward-specific state-value functions are formulated 
based on a correlation matrix to indicate the relative importance of the objectives on each other \citep{Zhan2019}. 
However, they need to tune this matrix weight to find the proper inter-objective relationship.
\citet{Abdolmaleki2020} develop a MO-RL framework 
based on the \textit{maximum a posteriori policy optimization} algorithm \citep{Abdolmaleki2018}.
They learned objective-specific policy distributions to find the Pareto solutions in a scale-invariant manner.
However, they still need to adjust objective-specific coefficients 
controlling the influence of objectives on the policy update.

In this paper, we propose a MO-RL algorithm for the continuous-valued state-action spaces without 
considering any preferences for different objectives.
In contrast to \citet{Abdolmaleki2020,Zhan2019,Chen2019}, a single-policy approach is devised which simplifies the algorithm architecture, and additionally there is no need for an initial assumption for reward preference.
As such, the devised algorithm can be considered scale-invariant.

The contributions of this paper are summarized as follows:

(1)  We devise a single-policy multi-objective RL algorithm without the importance of the competing objectives 
being as a priori knowledge.
We develop our algorithm  on the basis of Advantage Actor-Critic (A2C) \citep{A2C2012} 
using reward-specific state-value functions.

(2) We then provide a convergence analysis of the proposed scale-invariant MO-RL algorithm.

(3) We finally evaluate the devised algorithm over a multi-task problem.
Results show that it outperforms the single-objective A2C algorithm with a scalar reward
from the sample-efficiency and scale-invariance perspectives.

%The remainder of this paper is organized as follows: In Section II, some background requirements are explained. 
%In Section III, a multi-objective reinforcement algorithm is devised. 
%In Section IV, a convergence analysis is performed for the proposed algorithm.
%The experiment results are presented and discussed in Section V. 
%Finally, Section VI concludes the paper.

\textbf{Notations}: In this paper, we mainly use 
lower-case $a$ for scalars, 
bold-face lower-case $\textbf{a}$ for vectors and
bold-face uppercase $\textbf{A}$ for matrices. 
Further, $\textbf{A}^\top$ is the transpose of $\textbf{A}$,
$\|\textbf{a}\|$ and $\| \textbf{A} \|$ are the euclidean norm of $\textbf{a}$ 
and the corresponding induced matrix norm of $\textbf{A}$, respectively, 
and $\boldsymbol{\nabla}_{\textbf{a}} g(\cdot)$ %and $\boldsymbol{\nabla}^2_{\textbf{a}} g(\cdot)$ 
is the gradient vector %and Hessian matrix 
of multivariate function $g(\cdot)$ with respect to (w.r.t.) vector $\textbf{a}$. %, respectively.
We %show the components of a $n$-dimensional column vector $\boldsymbol{a}$ using the notation 
%$\boldsymbol{a}=[a_1,\ldots,a_n]^\top$ and 
indicate the m-th element of the vector $\boldsymbol{a}$ by $a_m$.
%and the $(n,m)$-th entry of matrix $\boldsymbol{A}$ by $[\boldsymbol{a}]_{n,m}$.
Further, $\{a_m\}_1^n$ collects the components of vector $\boldsymbol{a}$ from $m=1$ to $m=n$.
We use $\textbf{I}_n$, $\textbf{1}_n$, $\textbf{0}$ and $\textbf{e}_m$ to denote 
the identity matrix of size $n\times n$, 
a $n$-dimensional vector with all elements equal to one,
a vector with all elements equal to zero,
and a vector with all elements being zero except the $m$-th element that is one, respectively.

% /\/\/\/\/\/\/\/\/\/\/\/\/\/\/\/\/\/\/\\/\/\/\/\/\/\/\/\/\/\/\/\/\/\/\/\/\/\/\/\/\/\/\/\/\/\/\/\/\/\/\/\/\/\/\/\/\/\/\ %%
% /\/\/\/\/\/\/\/\/\/\/\/\/\/\/\/\/\/\/\\/\/\/\/\/\/\/\/\/\ MO-RL A2C  /\/\/\/\/\/\/\/\/\/\/\/\/\/\/\/\/\/\/\/\/\/\/\/\ %%
% /\/\/\/\/\/\/\/\/\/\/\/\/\/\/\/\/\/\/\\/\/\/\/\/\/\/\/\/\/\/\/\/\/\/\/\/\/\/\/\/\/\/\/\/\/\/\/\/\/\/\/\/\/\/\/\/\/\/\ %%
\section{Background}
\subsection{Multi-Objective Markov Decision Process}
A Multi-Objective Markov Decision Process (MO-MDP)  is expressed  according to  the tuple of
$\big( \mathcal{S},\mathcal{A},P_\mathcal{T}(\cdot), \{r_j(\cdot)\}_1^r \big)$,
where $\mathcal{S}$ is a set of states or the state space, 
$\mathcal{A}$ is a set of actions or the action space, 
$P_\mathcal{T}(\cdot): \mathcal{S} \times \mathcal{A} \times\mathcal{S} \to [0,~1]$ 
is the transition probability describing the system environment,
and $r_j(\cdot): \mathcal{S} \times \mathcal{A} \to \mathbb{R}$, for $j\in \{1,\ldots,r\}$, is the $j$-th immediate reward function.
The system state and action, at time $t$, are denoted by $\boldsymbol{s}_t \in \mathcal{S}$ and $\boldsymbol{a}_t \in \mathcal{A}$, respectively.
The transition probability $P_\mathcal{T}(\boldsymbol{s}_{t+1} | \boldsymbol{s}_t, \boldsymbol{a}_t)$ shows the probability that being in state $\boldsymbol{s}_t$ 
and performing action $\boldsymbol{a}_t$ leads to the next state $\boldsymbol{s}_{t+1}$. 
Therefore, we have: $\boldsymbol{s}_{t+1} \sim P_\mathcal{T}(\cdot | \boldsymbol{s}_t, \boldsymbol{a}_t)$.
The reward function $r_j(\boldsymbol{s}_t,\boldsymbol{a}_t)$ indicates the $j$-th immediate reward being obtained by transitioning from state $\boldsymbol{s}_t$ 
to state $\boldsymbol{s}_{t+1}$ by acting $\boldsymbol{a}_t$. 

In this work, we are interested in a stochastic policy representation.
For this, the action $\boldsymbol{a}_t$ is determined by drawing from a conditional policy distribution $\pi(\cdot | \boldsymbol{s}_t)$, i.e., $\boldsymbol{a}_t \sim \pi(\cdot | \boldsymbol{s}_t)$.
Based on the Markov property, the probability of a trajectory 
$\tau: \boldsymbol{s}_1 \to \boldsymbol{a}_1 \to \boldsymbol{s}_2 \to \boldsymbol{a}_2 \to \ldots \to \boldsymbol{s}_{T+1}$,
is determined by:
\begin{align}\label{EQ_markovianity}
	\mathsf{P}(\tau) :=& \mathsf{P}(\boldsymbol{s}_1,\boldsymbol{a}_1,\boldsymbol{s}_2,\boldsymbol{a}_2,\ldots,\boldsymbol{s}_{T+1})  
	= \mathsf{P}(\boldsymbol{s}_1) \prod_{t=1}^{T} \pi(\boldsymbol{a}_{t}|\boldsymbol{s}_{t}) P_\mathcal{T}(\boldsymbol{s}_{t+1} | \boldsymbol{a}_{t},\boldsymbol{s}_{t}).
\end{align}

In MO-MDP problems, the cumulative discounted rewards $\{R_j(t)\}_1^r$ are defined based on a summation over a finite horizon $T$ as:
$$
R_j(t) := \mathbb{E}\left\{ \sum_{k=t}^T \gamma^{k-t} r_j(\boldsymbol{s}_{k},\boldsymbol{a}_{k})  \right\},
$$
where the expectation is respect to $\mathsf{P}(\tau)$ and $\gamma: ~0 < \gamma \leq 1$ is the discount factor.
The aim is to find a single-agent stochastic policy, such that the cumulative discounted rewards $\{R_j(t)\}_1^r$ 
are maximized.
\begin{align}
	P_1: \qquad &\max_{ \pi(\cdot | \boldsymbol{s}_t) }~ ~~\big\{R_j(t)\big\}_1^r,~~ 0 \leq t \leq T \\
	&~{\rm s.t.} ~~ \boldsymbol{a}_t \sim \pi(\cdot | \boldsymbol{s}_t) \notag \\
	&~{\rm s.t.} ~~ \boldsymbol{s}_{t+1} \sim P_\mathcal{T}(\cdot|\boldsymbol{s}_t,\boldsymbol{a}_t). \notag 
\end{align}
Notice that $P_1$ is a multi-objective optimization problem and as such the maximization is regarded as the Pareto optimality perspective.

\subsection{A2C Algorithm}
Here, we address the structure of A2C \citep{A2C2012} as the basis of the multi-objective reinforcement algorithm 
we intend to devise.
For the conventional A2C algorithm, a single-objective MDP with a single immediate reward function 
$r(\cdot):\mathcal{S} \times\mathcal{A}  \to \mathbb{R}$ is taken into account, so $r=1$.
The aim is to design an optimal policy distribution being parameterized by a parameter ${\boldsymbol{\theta}} \in \Theta$,
where  $\Theta$ is a parameterization set of interest.
Here, we use the notation $\pi_{\boldsymbol{\theta}}(\cdot,\cdot)$ to show this parametric policy distribution.
Consequently, the trajectory probability (\ref{EQ_markovianity}) depends on ${\boldsymbol{\theta}}$ and can be expressed as:
\begin{align}\label{EQ_markovianity2}
	\mathsf{P}_{\boldsymbol{\theta}}(\tau) = \mathsf{P}(\boldsymbol{s}_1) \prod_{t=1}^{T} \pi_{\boldsymbol{\theta}}(\boldsymbol{a}_{t}|\boldsymbol{s}_{t}) P_\mathcal{T}(\boldsymbol{s}_{t+1} | \boldsymbol{a}_{t},\boldsymbol{s}_{t}).
\end{align}
Now, the objective can be expressed by $J({\boldsymbol{\theta}}) := \mathbb{E}_{\mathsf{P}_{\boldsymbol{\theta}}(\tau)} \big\{ \sum_{k=1}^T r(\boldsymbol{s}_k,\boldsymbol{a}_k) \big\}$ 
that is maximized with respect to policy distribution $\pi_{\boldsymbol{\theta}}(\cdot|\cdot)$.
Computing the gradient of $J({\boldsymbol{\theta}})$ gives:
\begingroup\makeatletter\def\f@size{10}\check@mathfonts
\begin{align*}
	&\nabla_{{\boldsymbol{\theta}}} J({\boldsymbol{\theta}}) \\
	~&\overset{a}=~ \mathbb{E}_{\mathsf{P}_{{\boldsymbol{\theta}}}(\tau)} \left\{ \sum_{k=1}^T r(\boldsymbol{s}_k,\boldsymbol{a}_k) \sum_{k=1}^T \nabla_{{\boldsymbol{\theta}}}\log \pi_{{\boldsymbol{\theta}}} (\boldsymbol{a}_k | \boldsymbol{s}_k) \right\} 
	\overset{b}= \sum_{k=1}^T \mathbb{E} \left\{ \nabla_{{\boldsymbol{\theta}}}\log \pi_{{\boldsymbol{\theta}}} (\boldsymbol{a}_k | \boldsymbol{s}_k) \sum_{k'=k}^T r(\boldsymbol{s}_{k'},\boldsymbol{a}_{k'}) \right\}  \\
	~&\overset{c}=~ \sum_{k=1}^T \mathbb{E}_{\boldsymbol{s}_k,\boldsymbol{a}_k} \Bigg\{ \nabla_{{\boldsymbol{\theta}}}\log \pi_{{\boldsymbol{\theta}}} (\boldsymbol{a}_k | \boldsymbol{s}_k) \underbrace{\mathbb{E}\bigg\{\sum_{k'=k}^T r(\boldsymbol{s}_{k'},\boldsymbol{a}_{k'}) \big| \boldsymbol{s}_k,\boldsymbol{a}_k \bigg\} }_{Q(\boldsymbol{s}_k,\boldsymbol{a}_k)} \Bigg\}  \\
	~&\overset{d}=~ \sum_{k=1}^T \mathbb{E}_{\boldsymbol{s}_k,\boldsymbol{a}_k} \Bigg\{ \nabla_{{\boldsymbol{\theta}}}\log \pi_{{\boldsymbol{\theta}}} (\boldsymbol{a}_k | \boldsymbol{s}_k) \Big(Q(\boldsymbol{s}_k,\boldsymbol{a}_k) - V(\boldsymbol{s}_k) \Big)  \Bigg\}  \\
	~&\overset{e}=~ \mathbb{E}\Bigg\{ \sum_{k=1}^T  \nabla_{{\boldsymbol{\theta}}}\log \pi_{{\boldsymbol{\theta}}} (\boldsymbol{a}_k | \boldsymbol{s}_k) \underbrace{\Big(r(\boldsymbol{s}_{k},\boldsymbol{a}_{k})+\gamma V(\boldsymbol{s}_{k+1}) - V(\boldsymbol{s}_k)\Big)}_{:=A(\boldsymbol{s}_k,\boldsymbol{a}_k)}   \Bigg\} \\
	~&=~ \mathbb{E}\Bigg\{ \sum_{k=1}^T  \nabla_{{\boldsymbol{\theta}}}\log \pi_{{\boldsymbol{\theta}}} (\boldsymbol{a}_k | \boldsymbol{s}_k) A(\boldsymbol{s}_k,\boldsymbol{a}_k)  \Bigg\},
\end{align*}
\endgroup
where $V(\cdot): \mathcal{S}\to \mathbb{R}$, $Q(\cdot,\cdot): \mathcal{S}\times \mathcal{A}\to \mathbb{R}$ 
and $A(\cdot,\cdot): \mathcal{S}\times \mathcal{A}\to \mathbb{R}$ are the state-value, action-value, and advantage functions, respectively.
For $(a)$ we used $\nabla_{{\boldsymbol{\theta}}} \mathsf{P}_{{\boldsymbol{\theta}}}(\tau) = \mathsf{P}_{{\boldsymbol{\theta}}}(\tau) ~\nabla_{{\boldsymbol{\theta}}} \log \mathsf{P}_{{\boldsymbol{\theta}}}(\tau)$, and $\nabla_{{\boldsymbol{\theta}}} \log \mathsf{P}_{{\boldsymbol{\theta}}}(\tau) = \sum_{k=1}^T \nabla_{{\boldsymbol{\theta}}} \log \pi_{{\boldsymbol{\theta}}} (\boldsymbol{a}_k|\boldsymbol{s}_k)$ 
based on Eq. (\ref{EQ_markovianity2}).
For (b) we considered the causality; the current action does not affect previous rewards, 
for (c) the definition of action-value function is applied,
for (d) we used the fact that including a bias term, here $V(\boldsymbol{s}_k)$, does not change the result due to 
$\mathbb{E}_{\boldsymbol{a}_k|\boldsymbol{s}_k} \left\{ \nabla_{\boldsymbol{\theta}} \log\pi_{\boldsymbol{\theta}}(\boldsymbol{a}_k|\boldsymbol{s}_k) \right\} = \boldsymbol{0} $
and for (e) the Bellman's equation $Q(\boldsymbol{s}_k,\boldsymbol{a}_k)= \mathbb{E}_{\boldsymbol{s}_{k+1} | \boldsymbol{s}_k,\boldsymbol{a}_k}\{r(\boldsymbol{s}_{k},\boldsymbol{a}_{k}) + \gamma V(\boldsymbol{s}_{k+1})\}$ is exploited. 
Note that the parameterized policy distribution $\pi_{\boldsymbol{\theta}}(\cdot | \cdot)$ is managed by an actor agent which 
can employ a neural network to generate action based on a given state.

Here, it is of benefit to remark on two practical points of the A2C algorithm.
First, a Stochastic Gradient Descent (SGD) is applied in A2C for which 
the parameter ${\boldsymbol{\theta}}$ is updated by the actor agent based on the following rule:
\begin{align*}
{\boldsymbol{\theta}} \leftarrow {\boldsymbol{\theta}} + \mu_a \nabla_{{\boldsymbol{\theta}}} \hat{J}({\boldsymbol{\theta}}),
\end{align*}
where $\mu_a$ is the actor learning rate and $\nabla_{{\boldsymbol{\theta}}} \hat{J}({\boldsymbol{\theta}})= \sum_{k=1}^T  \nabla_{{\boldsymbol{\theta}}}\log \pi_{{\boldsymbol{\theta}}} (\boldsymbol{a}_k | \boldsymbol{s}_k) A(\boldsymbol{s}_k,\boldsymbol{a}_k)$.
Additionally, it is conventional to represent the state-value function $V(\boldsymbol{s}_k)$ by a ${\boldsymbol{\phi}}$-parameterized approximation $V_{\boldsymbol{\phi}}(\boldsymbol{s}_k)$
with ${\boldsymbol{\phi}} \in \Phi$, where $\Phi$ is a parameterization set of interest. 
A neural network with parameter ${\boldsymbol{\phi}}$ can be employed by a critic agent for this representation.
Accordingly, the advantage function can be represented by $A_{{\boldsymbol{\phi}}}(\boldsymbol{s}_k,\boldsymbol{a}_k) :=  r(\boldsymbol{s}_{k},\boldsymbol{a}_{k})+\gamma V_{\boldsymbol{\phi}}(\boldsymbol{s}_{k+1}) -V_{\boldsymbol{\phi}}(\boldsymbol{s}_k)$.
Then, based on the Bellman's equation $V(\boldsymbol{s}_k) = \mathbb{E}_{\boldsymbol{s}_{k+1},\boldsymbol{a}_k | \boldsymbol{s}_k} \{ r(\boldsymbol{s}_{k},\boldsymbol{a}_{k})+\gamma V(\boldsymbol{s}_{k+1}) \}$,
the following objective, called critic loss, is considered to update parameter ${\boldsymbol{\phi}}$:
\begin{align*}
\sum_{k=1}^T \Big( \mathbb{E}_{\boldsymbol{s}_{k+1},\boldsymbol{a}_k | \boldsymbol{s}_k} \big\{ r(\boldsymbol{s}_{k},\boldsymbol{a}_{k})+\gamma V(\boldsymbol{s}_{k+1}) -V_{\boldsymbol{\phi}}(\boldsymbol{s}_k)  \big\} \Big)^2.
\end{align*}
However, in practice, the following SGD approach is leveraged:
\begin{align*}
{\boldsymbol{\phi}} \leftarrow {\boldsymbol{\phi}} + \mu_c  \sum_{k=1}^T A_{{\boldsymbol{\phi}}}(\boldsymbol{s}_k,\boldsymbol{a}_k) \: \nabla_{{\boldsymbol{\phi}} } V_{\boldsymbol{\phi}}(\boldsymbol{s}_k),
\end{align*}
where $\mu_c$ is the critic learning rate.
It is noteworthy that the update process of the actor agent can also be expressed based on the advantage function as:
\begin{align*}
{\boldsymbol{\theta}} \leftarrow {\boldsymbol{\theta}} + \mu_a \sum_{k=1}^T  \nabla_{{\boldsymbol{\theta}}}\log \pi_{{\boldsymbol{\theta}}} (\boldsymbol{a}_k | \boldsymbol{s}_k) A_{\boldsymbol{\phi}}(\boldsymbol{s}_{k},\boldsymbol{a}_{k}).
\end{align*}

\section{Multi-Objective A2C Algorithm}
Here, we devise a multi-objective RL approach on the grounds of A2C algorithm and the following Lemma \citep{Schafler2002,Pingchuan2020}:
\begin{lem}\label{Lemma_1}
	Assume a vector-valued multivariate function $\boldsymbol{f}=(f_1,\ldots,f_r),~ f_j: \mathbb{R}^n \to \mathbb{R}$
	for $j \in \{1,\ldots,r\}$.
	Define $\boldsymbol{q}(\cdot) = \sum_{j=1}^r \alpha^*_j \nabla f_j(\cdot)$, 
	then $-\boldsymbol{q}(\cdot)$ is a descent direction for all functions $\{f_j(\cdot)\}_1^r$, where $\{\alpha_j^* \}$ is the solution of following optimization problem:
	\begin{align*}
		Q_1: \qquad &\min_{ \{\alpha_j\}_1^r }~~~ \Big\| \sum_{j=1}^r \alpha_j \nabla f_j(\cdot) \Big\|^2,\quad {\rm s.t.} ~~ \sum_{j=1}^r \alpha_j = 1,~~ \alpha_j \geq 0 ~~~{\rm for}~ j\in\{1,\ldots,r\}. \notag 
	\end{align*}
\end{lem}
%\begin{proof}
%	See Theorem 2.1 of \cite{Schafler2002}.
%\end{proof}
Accordingly, we can get:
\begin{corollary}
	The solution of $Q_1$, for all $\alpha_j \geq 0$, reads:
	\begin{align}\label{EQ_alpha}
		\boldsymbol{\alpha}^* = \frac{ \left( \nabla F(\cdot)^\top \nabla F(\cdot) \right)^{-1} \boldsymbol{1}_r }
		{\boldsymbol{1}_r^{\top}\left( \nabla F(\cdot)^\top \nabla F(\cdot) \right)^{-1} \boldsymbol{1}_r},
	\end{align}
	where $\nabla F(\cdot)$ is an $n\times r$ matrix with $\nabla F(\cdot)= [\nabla f_1,\ldots,\nabla f_r](\cdot)$.
	Note that if there exists $l \in \{1,\ldots,r\}$ for which $\boldsymbol{e}_l^T \dfrac{ \left( \nabla F(\cdot)^\top \nabla F(\cdot) \right)^{-1} \boldsymbol{1}_r }
	{\boldsymbol{1}_r^{\top}\left( \nabla F(\cdot)^\top \nabla F(\cdot) \right)^{-1} \boldsymbol{1}_r} < 0$, then $\alpha^*_l=0$.
\end{corollary}

In the sequels, we develop a single-agent multi-objective A2C algorithm based on the result of Lemma \ref{Lemma_1}.
We call the proposed algorithm \textit{MO-A2C}.
For this, we formulate multi-objective actor (MO-actor) and multi-objective critic (MO-critic) agents as follows:
\subsection{\textbf{MO-Critic Agent}}
Consider a MO-MDP with immediate reward functions $\{r_j(\cdot)\}_1^r$. 
We formulate a MO-critic agent applying a shared ${\boldsymbol{\phi}}$-parameterized neural network 
in order to learn multiple state-value functions $\{V_{{\boldsymbol{\phi}},j}(\cdot)\}_1^r$ corresponding to rewards $\{r_j(\cdot)\}_1^r$. 
We thus introduce the $j$-th advantage function using:
$$
A_{{\boldsymbol{\phi}},j}(\boldsymbol{s}_t,\boldsymbol{a}_t) = r_j(\boldsymbol{s}_{t},\boldsymbol{a}_{t}) + \gamma V_{{\boldsymbol{\phi}},j}(\boldsymbol{s}_{t+1})-V_{{\boldsymbol{\phi}},j}(\boldsymbol{s}_t),~~{\rm for}~ j \in \{1,\ldots,r\}.
$$
Then, based on the Lemma \ref{Lemma_1}, we establish the following reward-specific MO-critic loss:
\begin{align}\label{EQ_Advantage}
	\hat{J}_{\rm moc,j}({\boldsymbol{\phi}}) = \sum_{k=1}^T A^2_{{\boldsymbol{\phi}},j}(\boldsymbol{s}_k,\boldsymbol{a}_k),
\end{align}	
with the MO-critic agent updating ${\boldsymbol{\phi}}$ by the rule:
\begin{align}\label{EQ_SGC_critic}
	{\boldsymbol{\phi}} \leftarrow {\boldsymbol{\phi}} - \mu_c \sum_{j=1}^r	\alpha_{\rm moc,j} \nabla_{{\boldsymbol{\phi}}} \hat{J}_{\rm moc,j}({\boldsymbol{\phi}}) ,
\end{align}	
where $\nabla_{{\boldsymbol{\phi}}} \hat{J}_{\rm moc,j}({\boldsymbol{\phi}}) = -\sum_{k=1}^T A_{{\boldsymbol{\phi}},j}(\boldsymbol{s}_k,\boldsymbol{a}_k) \nabla_{{\boldsymbol{\phi}}} V_{{\boldsymbol{\phi}},j}(\boldsymbol{s}_k)$ and $\{\alpha_{\rm moc,j}\}_1^r$ are obtained by:
\begin{align}\label{EQ_MO-criticupdate}
	\boldsymbol{\alpha}_{\rm moc} \:\:=\!\!  \operatorname*{argmin}_{\begin{small} \begin{array}{c} \{\alpha_j \geq 0\}_1^r \\ \sum_{j=1}^r \alpha_j=1	\end{array}		\end{small}} 
	\bigg\| \sum_{j=1}^r \alpha_j \: \nabla_{{\boldsymbol{\phi}}} \hat{J}_{\rm moc,j}({\boldsymbol{\phi}}) \bigg\|^2.
\end{align}

\subsection{\textbf{MO-Actor Agent}}
For the MO-actor agent, we consider a ${\boldsymbol{\theta}}$-parameterized neural network, which receives $\boldsymbol{s}_k$ 
and outputs the policy distribution $\pi_{{\boldsymbol{\theta}}}(\cdot|\boldsymbol{s}_k)$, from which the action vector $\boldsymbol{a}_k$ is drawn.
Accordingly, the following approach, which slightly differs from the MO-critic updating procedure, 
is devised for the MO-actor agent.
The $j$-th reward-specific loss is first created:
\begin{align}\label{EQ_actor_loss}
	\hat{J}_{\rm moa,j}({\boldsymbol{\theta}},{\boldsymbol{\phi}}) = -\sum_{k=1}^T  \log\pi_{{\boldsymbol{\theta}}}(\boldsymbol{a}_k|\boldsymbol{s}_k)   A_{{\boldsymbol{\phi}},j}(\boldsymbol{s}_k,\boldsymbol{a}_k).
\end{align}	
Then, the MO-actor agent updates ${\boldsymbol{\theta}}$ by the rule:
\begin{align}\label{EQ_SGC_actor}
	{\boldsymbol{\theta}} \leftarrow {\boldsymbol{\theta}} - \mu_a \sum_{j=1}^r	\alpha_{\rm moa,j} \nabla_{{\boldsymbol{\theta}}}\hat{J}_{\rm moa,j}({\boldsymbol{\theta}},{\boldsymbol{\phi}}) ,
\end{align}	
where $\{\alpha_{\rm moa,j}\}_1^r$ are found by:
\begin{align}\label{EQ_MOAupdate}
	\boldsymbol{\alpha}_{\rm moa} \:\:=\!\! \operatorname*{argmin}_{\begin{small} \begin{array}{c} \{\alpha_j \geq 0\}_1^r \\ \sum_{j=1}^r \alpha_{j}=1	\end{array}		\end{small}} 
	\bigg\| \sum_{j=1}^r \alpha_j \:\nabla_{{\boldsymbol{\theta}}} J_{\rm moa,j}({\boldsymbol{\theta}},{\boldsymbol{\phi}}) \bigg\|^2,
\end{align}
with $  J_{\rm moa,j}({\boldsymbol{\theta}},{\boldsymbol{\phi}}) =- \mathbb{E}\left\{\sum_{k=1}^T  \log\pi_{{\boldsymbol{\theta}}}(\boldsymbol{a}_k|\boldsymbol{s}_k)   A_{{\boldsymbol{\phi}},j}(\boldsymbol{s}_k,\boldsymbol{a}_k) \right\}$.
%	with $  J_{\rm moa,j}(\cdot) = \mathbb{E}\left\{ \hat{J}_{\rm moa,j}(\cdot)  \right\}$.
Note that, in contrast to MO-critic loss (see Eq. (\ref{EQ_MO-criticupdate})), 
we exploit the expected loss $\nabla_{{\boldsymbol{\theta}}} {J}_{\rm moa,j}({\boldsymbol{\theta}},{\boldsymbol{\phi}})$ to optimize $\boldsymbol{\alpha}_{\rm moa}$ in Eq. (\ref{EQ_MOAupdate}). 
To estimate it, we use a \textit{moving average} averaging of $\nabla_{{\boldsymbol{\theta}}} \hat{J}_{\rm moa,j}({\boldsymbol{\theta}},{\boldsymbol{\phi}})$ over different episodes and name it as \textbf{\textcolor{black}{episodic} average}.
\begin{figure}[t]
	\begin{center}
		\includegraphics[width=110 mm]{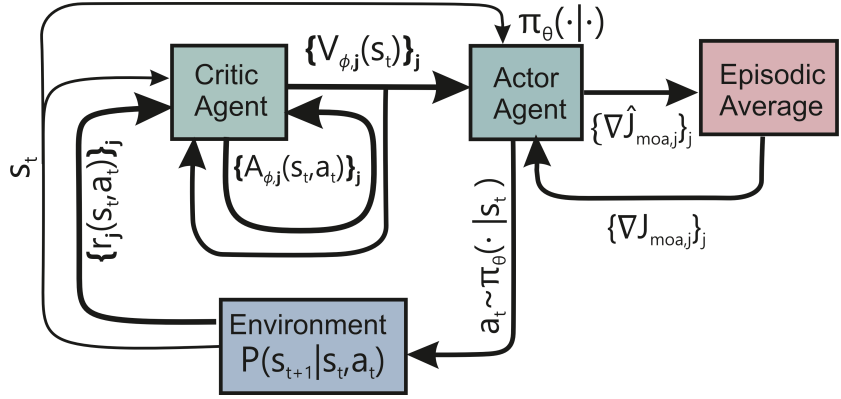}
		\vspace{5 pt}
		\caption{The diagram of proposed \textit{MO-A2C} algorithm. 
			A thick line shows reward-specific information flows
			while a thin line indicates a single information flow.  \label{Fig_MO_A2C}} 
	\end{center} 
\end{figure}

Figure \ref{Fig_MO_A2C} illustrates the diagram of devised multi-objective A2C algorithm.
The structure of this algorithm is also shown in Algorithm \ref{Alg1} 
with $E_{\rm max}$ being the number of episodic realizations the algorithm is learned for.

\begin{algorithm}[t]
	\caption{$~~$Pseudo-Code of \textit{MO-A2C}.}
	\label{Alg1}
	\small
	%\begin{algorithmic}[1]
		\For{$episode=1$ {\bfseries to} $E_{\max}$}{
		\textbf{Input:} Initial state vector $ \boldsymbol{s}_0$, MO-actor and MO-critic agents parameterized by $ {{\boldsymbol{\theta}}}$ and $ {{\boldsymbol{\phi}}}$.\\
		\For{$t=1$ {\bfseries to} $T$}{
		Select an action $ \boldsymbol{a}_t$ following $\pi_{ {{\boldsymbol{\theta}}}}(\cdot |  \boldsymbol{s}_t)$,
		interact with the environment.\\
		Observe new state $ \boldsymbol{s}_{t+1}$ and immediate rewards $\{r_j(\boldsymbol{s}_t,\boldsymbol{a}_t)\}_1^r$.\\
		Computes $\{A_{{\boldsymbol{\phi}},j}(\boldsymbol{s}_t,\boldsymbol{a}_t)\}_1^r$ using estimated Value-functions $\{V_{\boldsymbol{{\boldsymbol{\phi}}},j}( \boldsymbol{s}_t)\}_1^r$ and based on Eq. (\ref{EQ_Advantage}).\\
		Buffer  $\{V_{\boldsymbol{{\boldsymbol{\phi}}},j}( \boldsymbol{s}_t)\}_1^r$, $\{A_{\boldsymbol{{\boldsymbol{\phi}}},j}( \boldsymbol{s}_{t},\boldsymbol{a}_t)\}_1^r$, $\{r_j(\boldsymbol{s}_t,\boldsymbol{a}_t)\}_1^r$ and $\log\big(\pi_{ {{\boldsymbol{\theta}}}}( \boldsymbol{a}_{t}| \boldsymbol{s}_t)\big)$. \\
		Compute episodic average of MO-actor losses $\{\nabla \hat{J}_{\rm moa,j}\}_1^r$ to estimate expected losses $\{\nabla {J}_{\rm moa,j}\}_1^r$.\\
		\If{update is needed,}{
		\vspace{6 pt}
		\textbf{MO-critic update process}: \\
		Obtain $\boldsymbol{\alpha}_{\rm moc}$ based on Eq. (\ref{EQ_MO-criticupdate}).\\
		Compute MO-critic losses $\{\hat{J}_{\rm moc,j}({\boldsymbol{\phi}})\}_1^r$ 
		and apply the rule:\\
		\vspace{-6 pt}
		$$
		 {\boldsymbol{\phi}} \leftarrow {\boldsymbol{\phi}} - \mu_c \sum_{j=1}^r	\alpha_{\rm moc,j} \nabla_{{\boldsymbol{\phi}}}\hat{J}_{\rm moa,j}({\boldsymbol{\phi}}).
		$$
		\vspace{-9 pt}\\
		\textbf{MO-actor update process}: \\
		Obtain $\boldsymbol{\alpha}_{\rm moa}$ based on Eq. (\ref{EQ_MOAupdate}).\\
		Eestimate $\{\nabla {J}_{\rm moa,j}\}_1^r$ by the \textbf{episodic averaging}.\\
		Compute MO-actor losses $\{ \hat{J}_{\rm moa,j}({\boldsymbol{\theta}},{\boldsymbol{\phi}})\}_1^r$ 
		and apply the rule: \\
		\vspace{-6 pt}
		$$
		{\boldsymbol{\theta}} \leftarrow {\boldsymbol{\theta}} - \mu_a \sum_{j=1}^r	\alpha_{\rm moa,j} \nabla_{{\boldsymbol{\theta}}}\hat{J}_{\rm moa,j}({\boldsymbol{\theta}},{\boldsymbol{\phi}}).
		$$
	}
	}
	}
	%\end{algorithmic}
\end{algorithm}

\section{Convergence Analysis of MO-A2C Algorithm}
Here, we intend to analyze the convergence of the proposed \textit{MO-A2C} algorithm.
For this, we first make some assumptions:

\textbf{Assumption 1:}
The following relation between the advantage functions $A_j(\cdot,\cdot)$ 
and their parametric representations $A_{\boldsymbol{\phi},j}(\cdot,\cdot)$ exists:
$$
A_{{\boldsymbol{\phi}},j}(\boldsymbol{s}_k,\boldsymbol{a}_k) = A_{j}(\boldsymbol{s}_k,\boldsymbol{a}_k) + \delta_{{\boldsymbol{\phi}},j}(\boldsymbol{s}_k),\qquad {\rm for}~ j\in\{1,\ldots,r\},
$$
which indicates that the difference between the approximated and exact values of the advantage function 
can be expressed based on an \textit{action-independent} drift function $\delta_{{\boldsymbol{\phi}},j}(\cdot)$.
\textcolor{black}{Although, this assumption might not be valid in the early stages of the algorithm,
	this can hold after some iterations.}
According to this assumption and Eq. (\ref{EQ_actor_loss}) we get:
\begin{align*}
	\mathbb{E} \left\{ \nabla_{{\boldsymbol{\theta}}} \hat{J}_{\rm moa,j}({\boldsymbol{\theta}},{\boldsymbol{\phi}})  \:\big|\: {\boldsymbol{\theta}},{\boldsymbol{\phi}} \right\} &= \nabla_{{\boldsymbol{\theta}}} {J}_{\rm moa,j}({\boldsymbol{\theta}},{\boldsymbol{\phi}}) 
	\overset{a}=- \mathbb{E}\left\{\sum_{k=1}^T  \nabla_{\boldsymbol{\theta}}\log\pi_{{\boldsymbol{\theta}}}(\boldsymbol{a}_k|\boldsymbol{s}_k)   A_{j}(\boldsymbol{s}_k,\boldsymbol{a}_k)\right\} \\
	& :=  \nabla_{{\boldsymbol{\theta}}} {J}_{\rm moa,j}({\boldsymbol{\theta}}),
\end{align*}
where for (a), we used $\mathbb{E}_{\boldsymbol{a}_k|\boldsymbol{s}_k} \left\{ \nabla_{\boldsymbol{\theta}} \log\pi_{\boldsymbol{\theta}}(\boldsymbol{a}_k|\boldsymbol{s}_k) \delta_{{\boldsymbol{\phi}},j}(\boldsymbol{s}_k) \right\} = \boldsymbol{0} $.

Additionally, we take into account two conventional assumptions 
based on the literature \citep{Qiu2021}.

\textbf{Assumption 2:} The MO-actor expected losses $\{{J}_{moa,j}({\boldsymbol{\theta}},{\boldsymbol{\phi}})\}_1^r$ are strongly convex with parameter $\gamma$ w.r.t ${\boldsymbol{\theta}}$. 
As such we get:
\begin{align*}
	{J}_{moa,j}({\boldsymbol{\theta}}',{\boldsymbol{\phi}}) - {J}_{moa,j}({\boldsymbol{\theta}},{\boldsymbol{\phi}}) \geq \nabla_{\boldsymbol{\theta}} {J}_{moa,j}({\boldsymbol{\theta}})^\top ({\boldsymbol{\theta}}'-{\boldsymbol{\theta}})+
	\frac{\gamma}{2}\|{\boldsymbol{\theta}}'-{\boldsymbol{\theta}} \|^2.
\end{align*}
Furthermore, they are Lipschitz smooth functions with constant $L$  w.r.t ${\boldsymbol{\theta}}$, so we have:
\begin{align*}
	{J}_{moa,j}({\boldsymbol{\theta}}',{\boldsymbol{\phi}}) - {J}_{moa,j}({\boldsymbol{\theta}},{\boldsymbol{\phi}}) \leq \nabla_{\boldsymbol{\theta}} {J}_{moa,j}({\boldsymbol{\theta}})^\top ({\boldsymbol{\theta}}'-{\boldsymbol{\theta}} )+
	\frac{L}{2}\|{\boldsymbol{\theta}}'-{\boldsymbol{\theta}} \|^2.
\end{align*}
Notice that Assumptions 2 is made for the expected losses $\{{J}_{moa,j}({\boldsymbol{\theta}},{\boldsymbol{\phi}})\}_1^r$ 
and not for the stochastic losses $\{\hat{J}_{moa,j}({\boldsymbol{\theta}},{\boldsymbol{\phi}})\}_1^r$.

\textbf{Assumption 3:} Consider the Jacobian matrix 
$\nabla {\boldsymbol{\hat{J}}}({\boldsymbol{\theta}},{\boldsymbol{\phi}})= [\nabla_{{\boldsymbol{\theta}}} \hat{J}_{\rm moa,1},\ldots,\nabla_{{\boldsymbol{\theta}}} \hat{J}_{\rm moa,r}]({\boldsymbol{\theta}},{\boldsymbol{\phi}}) $ where $\nabla\boldsymbol{J}({\boldsymbol{\theta}}) = \mathbb{E}\left\{  \nabla {\boldsymbol{\hat{J}}}({\boldsymbol{\theta}},{\boldsymbol{\phi}})  \:\big|\: {\boldsymbol{\theta}},{\boldsymbol{\phi}}  \right\}$.
Then, its conditional covariance is bounded by a positive semi-definite matrix $\bf{B}$:
$$
\mathbb{E}\left\{  \nabla {\boldsymbol{\hat{J}}}({\boldsymbol{\theta}},{\boldsymbol{\phi}})^\top \nabla {\boldsymbol{\hat{J}}}({\boldsymbol{\theta}},{\boldsymbol{\phi}}) \:\big|\: {\boldsymbol{\theta}},{\boldsymbol{\phi}} \right\} - \nabla {\boldsymbol{J}}({\boldsymbol{\theta}})^\top \nabla {\boldsymbol{J}}({\boldsymbol{\theta}}) \leq \bf{B}.
$$
This assumption indicates that the covariance matrix of Jacobian of the stochastic losses is upper-bounded by $\bf{B}$.
%\textbf{Assumption 4:} We assume that the error of the MO-critic agent is negligible such that for the advantage function, we get:
%$ A_{{\boldsymbol{\phi}},j} = r_j(\boldsymbol{s}_{t+1},\boldsymbol{a}_{t+1}) + \gamma V_{j}(\boldsymbol{s}_{t+1})-V_{j}(\boldsymbol{s}_t) $

Now, we have the following theorem for the \textit{MO-A2C} algorithm:
\begin{thm}\label{Thm1}
	Assume MO Problem $P_1$ with a ${\boldsymbol{\theta}}$-parameterized policy distribution $\pi_{\boldsymbol{\theta}}(\cdot|\cdot)$
	being optimized by SGDes  (\ref{EQ_SGC_critic}) and  (\ref{EQ_SGC_actor})
	with generated sequences $\{ {\boldsymbol{\phi}}^i \}_{i\in\mathbb{N}}$ and $\{ {\boldsymbol{\theta}}^i \}_{i\in\mathbb{N}}$
	and MO-actor learning rate $\mu_i \leq \min\left\{ \frac{1}{L},\frac{1}{L \| \bf{B}\|} \mathbb{E}_{\boldsymbol{\theta}}\Big\{  \dfrac{1}{\boldsymbol{1}_r^{\top}\left( \nabla \boldsymbol{J}({\boldsymbol{\theta}})^\top \nabla \boldsymbol{J}({\boldsymbol{\theta}}) \right)^{-1} \boldsymbol{1}_r} \Big\} \right\}$.
	Moreover, consider MO-actor expected losses $\{{J}_{moa,j}(\cdot)\}_1^r$ 
	and stochastic losses $\{\hat{J}_{moa,j}(\cdot)\}_1^r$  complying with Assumptions 2 and 3, respectively,
	and that there exists an optimal Pareto solution ${\boldsymbol{\theta}}^*$ of $P_1$,
	then we get:
	$$
	\mathbb{E} \|{\boldsymbol{\theta}}^{i+1} - {\boldsymbol{\theta}}^* \|  \leq (1-\gamma\mu_i) \:  \mathbb{E} \|{\boldsymbol{\theta}}^{i} - {\boldsymbol{\theta}}^* \| + 2\mu_i^2 \| \bf{B}\|.
	$$
\end{thm}
\begin{proof}
	Refer to the Supplementary Material.
\end{proof}

Accordingly, we can get:
\begin{corollary}\label{Cor3}
	Consider the SGD approach (\ref{EQ_SGC_actor}) with generated sequence $\{{\boldsymbol{\theta}}^n\}_{n\in\mathbb{N}}$ 
	and MO-actor learning rate $\{\mu_n\}_{n\in\mathbb{N}}$ complying with assumptions of Theorem \ref{Thm1}. 
	Set $\mu_n$ so that $\lim_{n\to\infty} \mu_n = 0$, then 
	$\lim_{n\to\infty} \mathbb{E}\| {\boldsymbol{\theta}}^{n} - {\boldsymbol{\theta}}^* \|^2 = 0$.
\end{corollary}
\begin{proof}
	We use the result of \citet{Turinici2021}.
	As such, based on Theorem \ref{Thm1} we have:
	\begin{align*}
		d_{n+1} - \epsilon \:&\leq\: (1-\gamma \mu_n)(d_n-\epsilon) - \mu_n(\gamma\epsilon-2\mu_n \| \bf{B}\|)\\
		\:&\overset{a}\leq\:  (1-\gamma \mu_n)(d_n-\epsilon) .
	\end{align*}
	where $d_n = \mathbb{E} \| {\boldsymbol{\theta}}^n - {\boldsymbol{\theta}}^* \|^2$ and $\epsilon > 0 $.
	For (a), we considered that $\gamma\epsilon-2\mu_n \| \bf{B}\| \geq 0$ for large $n$.
	Hence, for $\gamma \mu_n \leq 1$ it reads:
	$$
	[d_{n+1} - \epsilon]^+ \leq (1-\gamma \mu_n)[d_n-\epsilon]^+,
	$$	
	where $[x]^+ = x  + |x|$. By iterating, we get:
	$$
	[d_{n+k} - \epsilon]^+ \leq \prod_{i=0}^{k-1}(1-\gamma \mu_{n+i})\: [d_n-\epsilon]^+.
	$$
	Considering that $\lim_{k\to \infty} \prod_{i=0}^{k-1}(1-\gamma \mu_{n+i}) =0$, we have: $\lim_{m\to \infty} [d_m-\epsilon]^+$. Since it holds for any value $\epsilon > 0$, the statement follows.
\end{proof}
\textcolor{black}{Note that the result of Corollary \ref{Cor3} guarantees that a convergence-in-mean can be achieved 
	by choosing a suitable MO-actor learning rate.}

\section{Experiment}
\subsection{Evaluation over a Mlti-Task Problem}\label{Sec_Exp}
To empirically evaluate the devised \textit{MO-A2C} algorithm, 
we consider a practical multi-task problem from the context of edge caching for cellular networks.
For this, we take into account the system model presented in \citet{Ours2021}.
The environment of this problem is a mobile cellular network which serves requesting mobile users by employing base-stations.
The cellular network  operates in a time-slotted fashion with time index $t \in \{1,\ldots,T\}$, 
where $T$ is the total duration within which the network operation is considered.

The network itself constitutes file-requesting users, transmitting base-stations 
as well as a library containing $N$ files from which the users request files.
The users are spatially distributed across the network.
They are interested in different files based on a \textit{file popularity}
which determines the probability that a file is preferred by a typical user. 
As such, files will be requested with different probabilities.
The base-stations are also spatially distributed and are equipped with caches.
The base-station caches are with a limited capacity of $M$ files.
They proactively store files from the library at their caches.
For this, a probabilistic approach is used to place the files at base-station caches;
file $n \in \{1,\ldots,N\}$ is cached at a typical base-station in time $t$ with probability $p_{t,n}$.

To model the location of users and base-stations, 
we use two Poisson point processes with intensities $\lambda_u$ and $\lambda_b$, respectively.
The network employs the base-stations to multicast the cached files to satisfy users.
Moreover, the base-stations exploit a resource allocation to transmit different files.
As such, disjoint file-specific radio resources are allocated so that
file $n$ is transmitted at time $t$ by occupying bandwidth $w_{t,n}$.
The total amount of radio resources being needed to satisfy users causes a network load, 
namely as \textbf{bandwidth consumption cost}.

Not all users can be successfully served by the network due to a reception outage probability.
As such, at time $t$, a typical user preferring file $n$ is not able to receive the needed file with 
a reception outage probability $\mathcal{O}_{t,n}$.
The unsatisfied users will fetch the file directly from the network using a reactive transmission 
and consuming a sufficient amount of radio resources.
However, this causes a network load, namely as \textbf{backhaul cost}, 
due to on-demand file transmission from the core-network to the requesting user.

The aim of this problem is to design a cache policy in order to satisfy as many users as possible 
with the minimum level of resource consumption and backhaul cost.
More specifically, the cache policy should consider three competing objectives as follows.
A \textbf{quality-of-service (QoS) metric} that measures the percentage of users 
that can successfully receive their needed files.
This metric shows the probability that a requesting user is properly satisfied by downloading its needed file, 
and it can be expressed at time $t$ based on the file-specific reception outage probability $\mathcal{O}_{t,n}$.
The other objective is a \textbf{bandwidth (BW) consumption metric}
indicating the total radio resources being allocated to respond users.
We finally consider a \textbf{backhaul (BH) cost} measuring the network load needed 
to fetch files directly from the network than the cache of base-stations.
Note that we express QoS metric, BW consumption, and BH load objectives at time $t$ 
based on negative immediate rewards 
that are denoted by $r_{\rm QoS}(t)$, $r_{\rm BW}(t)$ and $r_{\rm BH}(t)$, respectively.
Note that these immediate rewards depend on the file-specific cache probabilities $\{p_{t,n}\}_1^N$ 
and file-specific resource allocation $\{w_{t,n}\}_1^N$ \citep{Ours2021}.
As such, a multi-task problem with three competing objectives can be formulated.

For this problem, we define the system state $\boldsymbol{s}_t$ as a vector containing 
file-specific request probabilities of users from the network. 
We denote this vector by $r_{t,n}$ which shows 
the probability that file $n$ is requested from the network by a typical user at time $t$.
The system action $\boldsymbol{a}_t$ is a vector comprising of the file-specific cache probabilities $p_{t,n}$ 
and file-specific resource allocation $w_{t,n}$.
Hence, the state and action vectors are expressed by:
$$
\boldsymbol{s}_t = \big[\{r_{t,n}\}_{1}^N\big]^\top,\quad \boldsymbol{a}_t= \big[\{p_{t,n}\}_{1}^N,\:\{w_{t,n}\}_{1}^N\big]^\top, \qquad~ \mbox{for} ~~t\in \{1,\ldots,T\}.
$$

This cache policy problem can be formulated based on a MO-MDP with a continuous state-action space \citep{Ours2021}, 
and as such it can be designed
based on a multi-objective RL algorithm.
Therefore, we apply the algorithm \ref{Alg1} for this problem whose solution is denoted by \textit{MO-A2C}.
We also construct a scalar reward based on a linear combination and then use a single-objective A2C algorithm 
whose solution is denoted by \textit{SO-A2C}.

\subsection{Experiment Setup and Hyper-parameters}
We consider the following settings for the considered system environment.
The number of files is $N=100$, the capacity of base-stations $M=10$,
and the spatial intensity of base-stations and users are $\lambda_b=10$ 
and $\lambda_u=10^5$, respectively, with the units of points/km$^2$.
The desired rate of transmission is $1$ Mbits/second. 
This quantity affects the reception outage probability.
The total number of time-slots is $T$ = 256 and the discount factor is set $\gamma$ = 0.96.

For the \textit{MO-A2C} algorithm, the actor and critic learning rates are set to $1\times10^{-3}$.
Two separate neural networks, each with one hidden layer, represent the MO-actor and MO-critic agents. 
The MO-critic network outputs three values representing 
the reward-specific state-value functions $V_{{\boldsymbol{\phi}},j}(\cdot)$.
The number of neurons in the hidden layer for the critic is $64$
and the rectified linear unit (ReLU) activation function is used for its neuron.
The MO-actor network represents the RL single-policy distribution $\pi_{{\boldsymbol{\theta}}}(\cdot,\cdot)$.
The number of neurons in the hidden layer for the actor network is $128$.

For the \textit{SO-A2C} algorithm, the actor and critic learning rates are the same as \textit{MO-A2C}.
Moreover, the same architecture is considered for the actor and critic neural networks,
except that the number of neurons in the hidden layer of the actor is $64$, 
which is set to give the best performance.
The critic network outputs only a single state-value function related to the scalar reward.
Notice that the scalar reward is obtained based on a linear combination of aforementioned rewards $r_{\rm Qos}(t)$, $r_{\rm BW}(t)$ and $r_{\rm BH}(t)$ as follows:
$$
r_{\rm sc} = \lambda_{\rm QoS}\: r_{\rm QoS}(t)+\lambda_{\rm BW}\:r_{\rm BW}(t)+\lambda_{\rm BH}\:r_{\rm BH}(t),
$$
where $\lambda_{\rm QoS}$, $\lambda_{\rm BW}$ and $\lambda_{\rm BH}$ 
are the scalarization scales.
We evaluate the following combinations for these scales: 
$$
[\lambda_{\rm QoS}, \:\lambda_{\rm BW},\: \lambda_{\rm BH}] \in \big\{[1,\:1,\:1],\:[0.1,\:1,\:1],\:[1,\:10,\:0.1]\big\}.
$$

\subsection{Experiment Results}
We apply the multi-task problem explained in Section \ref{Sec_Exp}.
Additionally, it is noteworthy that the considered rewards, i.e., $r_{\rm Qos}(t)$, $r_{\rm BW}(t)$ and $r_{\rm BH}(t)$, 
are re-scaled to lie in the range $[0,\:1]$ and then are used by \textit{MO-A2C} and \textit{SO-A2C} algorithms.

\begin{figure}
	\centering
	\begin{minipage}{.5\textwidth}
		\centering
		\hspace{-20 pt}\includegraphics[width=7 cm]{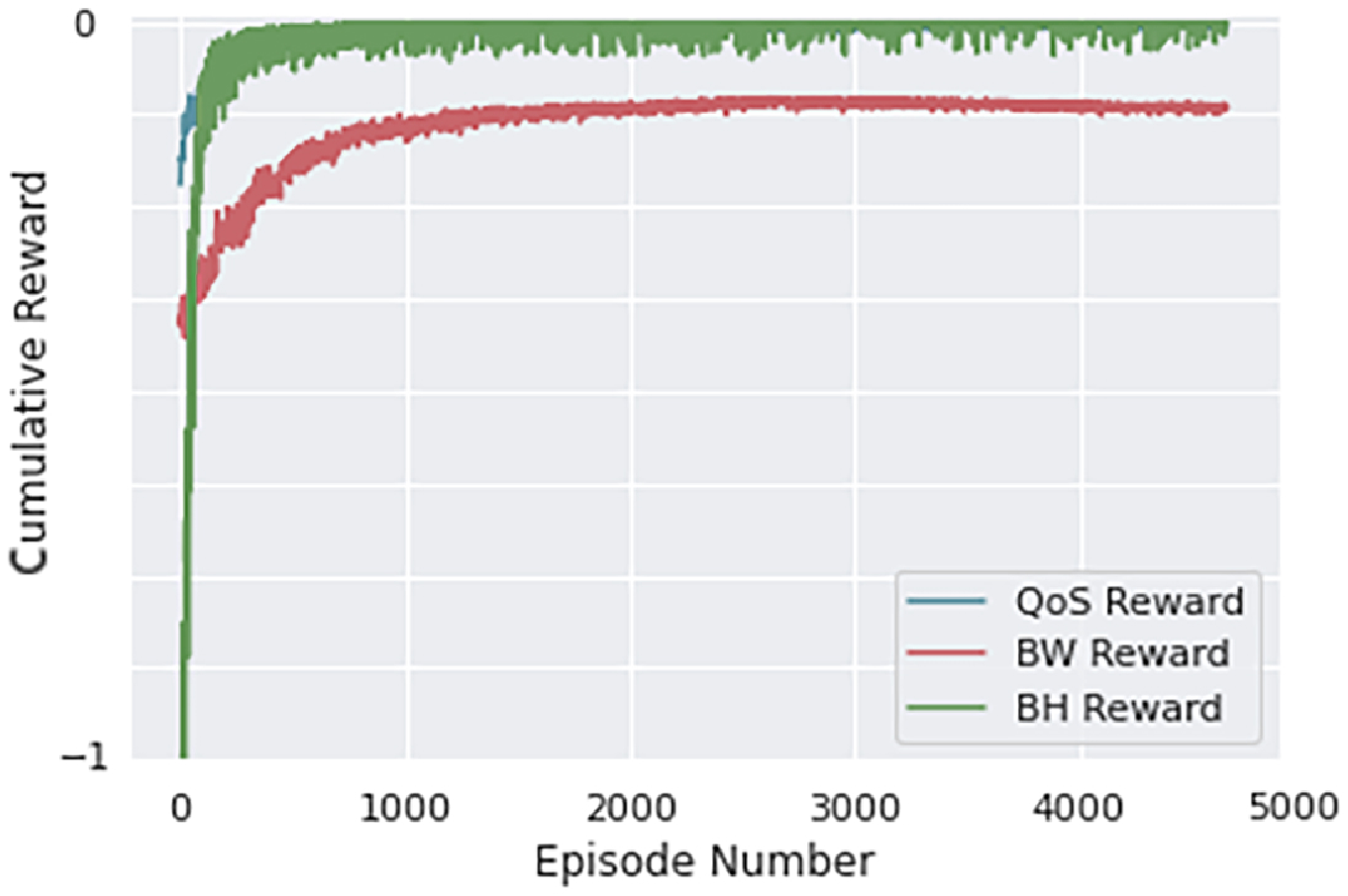}
		\captionof{figure}{Discounted cumulative reward for QoS \\metric, BW  consumption and BH load obtained by \\ \textit{MO-A2C}.}
		\label{Fig_MOA2C_SA}
	\end{minipage}%
	\begin{minipage}{.5\textwidth}
		\centering
		\hspace{-15 pt}\includegraphics[width=7 cm]{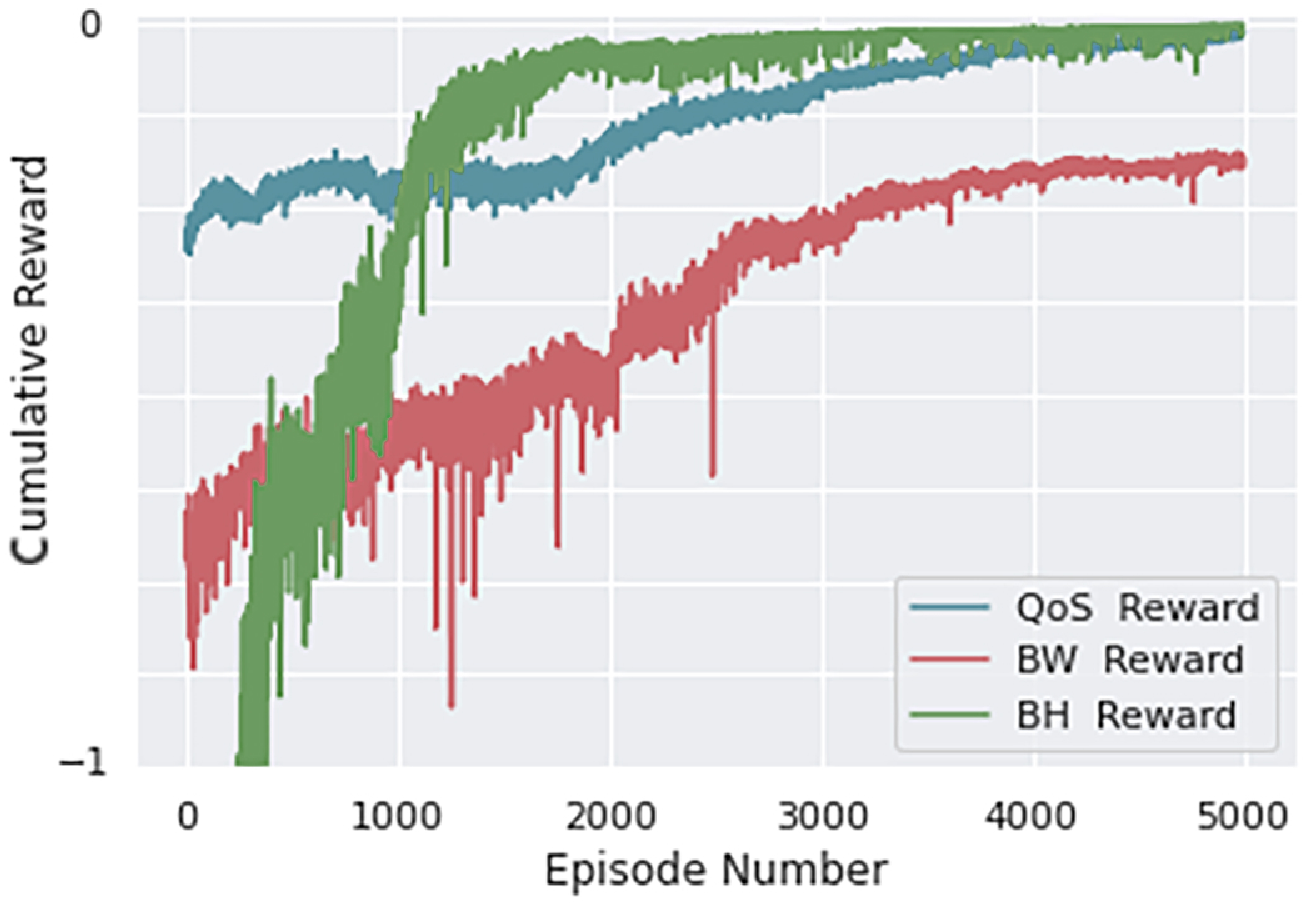}
		\captionof{figure}{Discounted cumulative reward for QoS \\metric, BW consumption and BH load obtained by \\ \textit{SO-A2C}.}
		\label{Fig_SOA2C_SA}
	\end{minipage}
\end{figure}

We evaluate the sample efficiency of \textit{MO-A2C} 
and compare it to \textit{SO-A2C} with scalarization scales $[\lambda_{\rm QoS}, \:\lambda_{\rm BW},\: \lambda_{\rm BH}] = \boldsymbol{1}_3$.
The training performance of \textit{MO-A2C} and \textit{SO-A2C} are plotted in Figures \ref{Fig_MOA2C_SA} and \ref{Fig_SOA2C_SA},
in terms of the cumulative rewards of mentioned metrics (QoS metric, BW  consumption, and BH load) for different episodes ($E_{\rm max}$).
%Note that these rewards are first re-scaled to lie in the range $[0,\:1]$ 
%and then are used by \textit{MO-A2C} and \textit{SO-A2C} algorithms.
According to Figures \ref{Fig_MOA2C_SA} and \ref{Fig_SOA2C_SA}, 
it can be seen that \textit{MO-A2C} outperforms \textit{SO-A2C} from sample-efficiency perspective, 
as it can be learned after $2\times 10^3$ episodes while \textit{SO-A2C} needs more that $5\times 10^3$ episode samples to be learned.

\begin{figure}
	\centering
	\begin{minipage}{.5\textwidth}
		\centering
		\hspace{-20 pt}\includegraphics[width=7 cm]{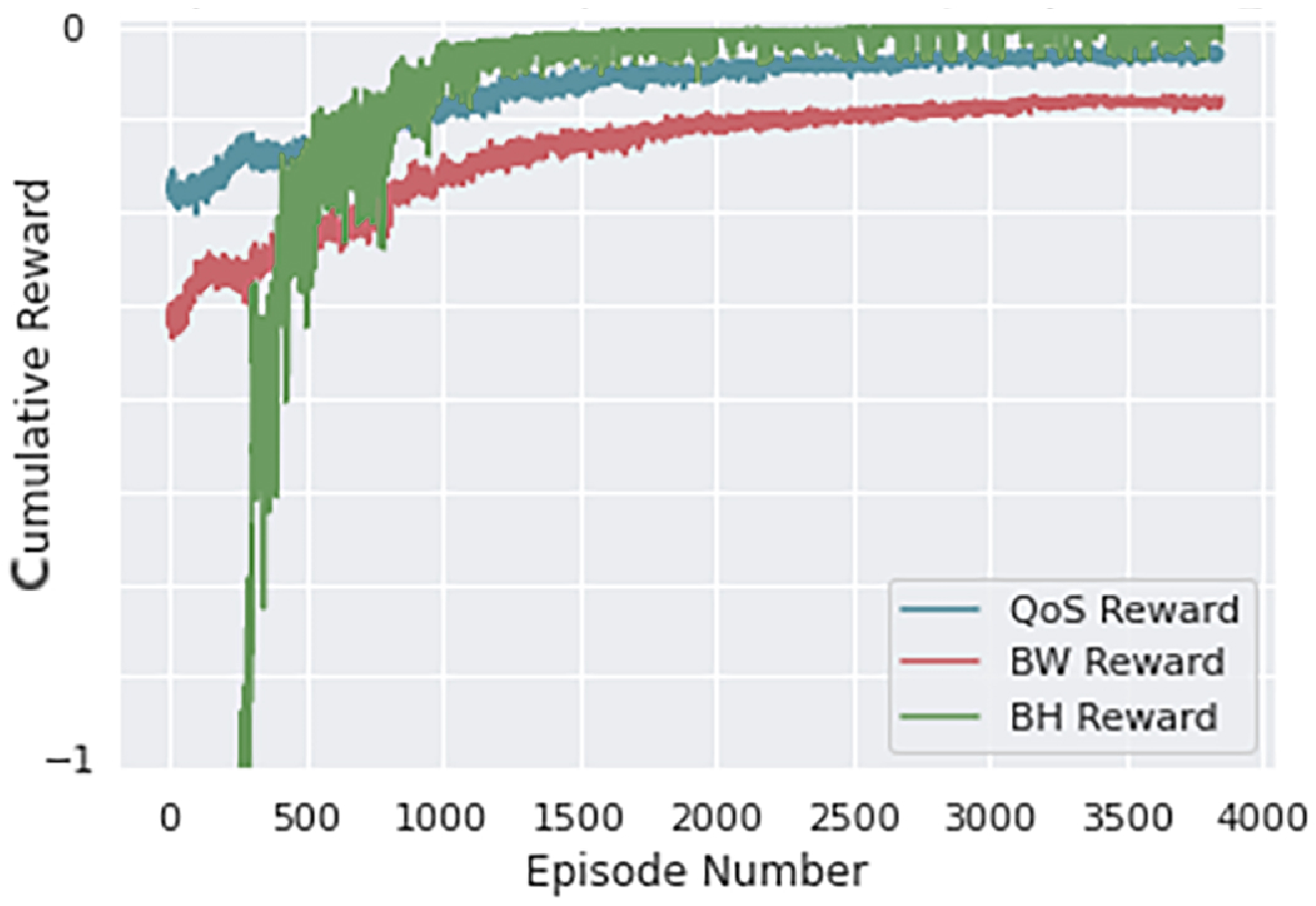}
		\captionof{figure}{Discounted cumulative reward for QoS \\metric, BW  consumption and BH load obtained by \\ \textit{MO-A2C}.
			QoS reward is multiplied by $0.1$ and \\other rewards are kept unchanged.}
		\label{Fig_MOA2C_SD}
	\end{minipage}%
	\begin{minipage}{.5\textwidth}
		\centering
		\hspace{-15 pt}\includegraphics[width=7 cm]{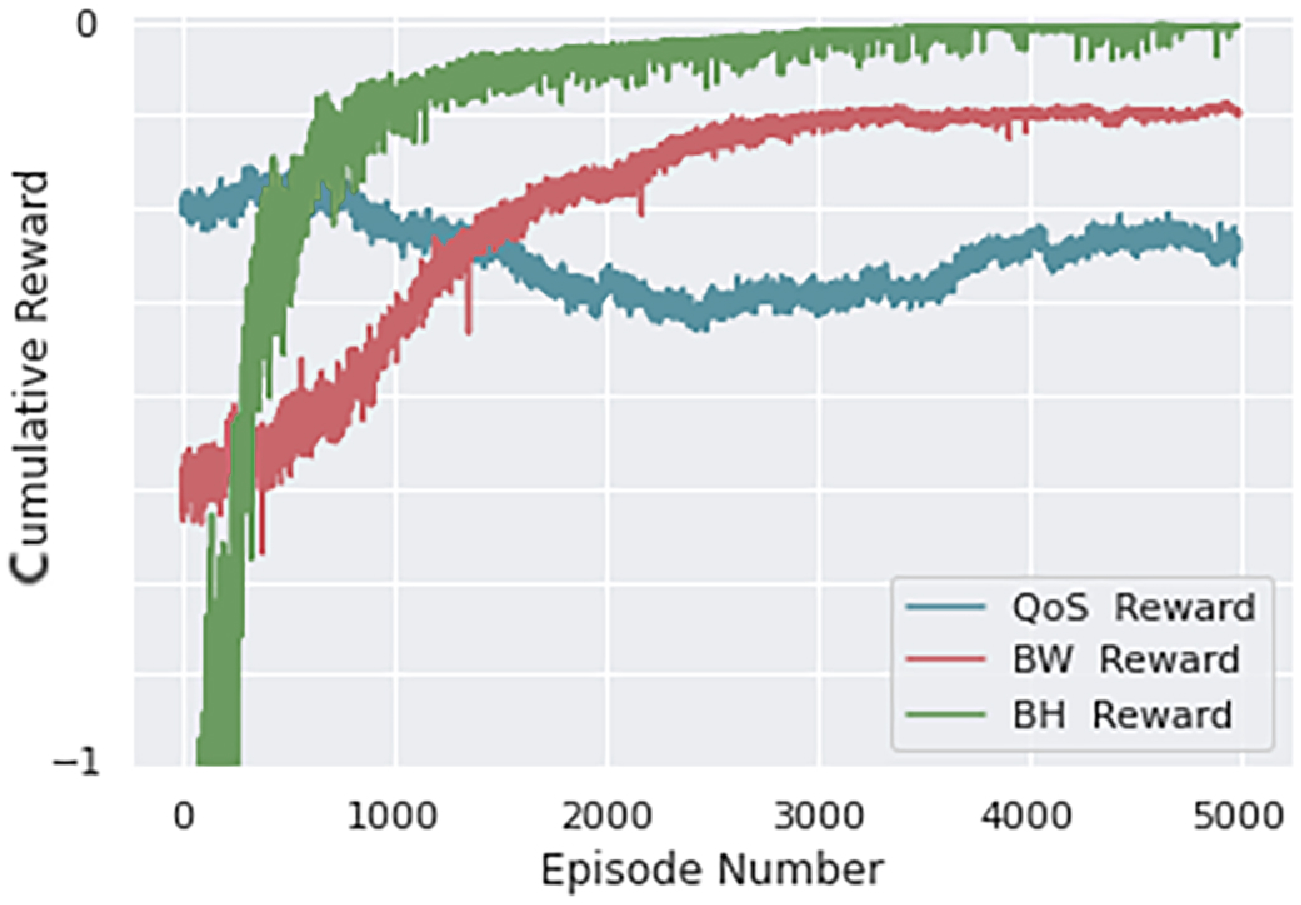}
		\captionof{figure}{Discounted cumulative reward for QoS \\metric, BW consumption and BH load obtained by \\ \textit{SO-A2C}.
			To constitute a scalarized reward, we use \\the scales $[\lambda_{\rm QoS},\: \lambda_{\rm BW},\: \lambda_{\rm BH}] = [0.1,\:1,\:1]$.}
		\label{Fig_SOA2C_SD}
	\end{minipage}
\end{figure}

To evaluate the scale-dependency of \textit{MO-A2C} and \textit{SO-A2C}, we consider a test scenario.
For \textit{SO-A2C}, we use the scalarization scales $[\lambda_{\rm QoS}, \:\lambda_{\rm BW},\: \lambda_{\rm BH}] = [0.1,\:1,\:1]$.
For \textit{MO-A2C}, we multiply the QoS reward by $0.1$ and keep the other two rewards
unchanged, and then evaluate the algorithms over the re-scaled rewards.
For this scenario, we plot the training performance of \textit{MO-A2C} and \textit{SO-A2C} 
in Figures \ref{Fig_MOA2C_SD} and \ref{Fig_SOA2C_SD},
in terms of the cumulative rewards for different episodes.
Based on Figures \ref{Fig_MOA2C_SD} and \ref{Fig_SOA2C_SD}, it can be inferred that 
\textit{MO-A2C} is more insensitive towards the scaling than \textit{SO-A2C}.
More specifically, \textit{MO-A2C} has been able to learn the RL agent after around $3\times 10^3$ episodes despite of QoS reward 
being re-scaled.
However, \textit{SO-A2C} agent has not been properly learned even after $5\times 10^3$ episodes.

\begin{figure}
	\centering
	\begin{minipage}{.5\textwidth}
		\centering
		\hspace{-20 pt}\includegraphics[width=7 cm]{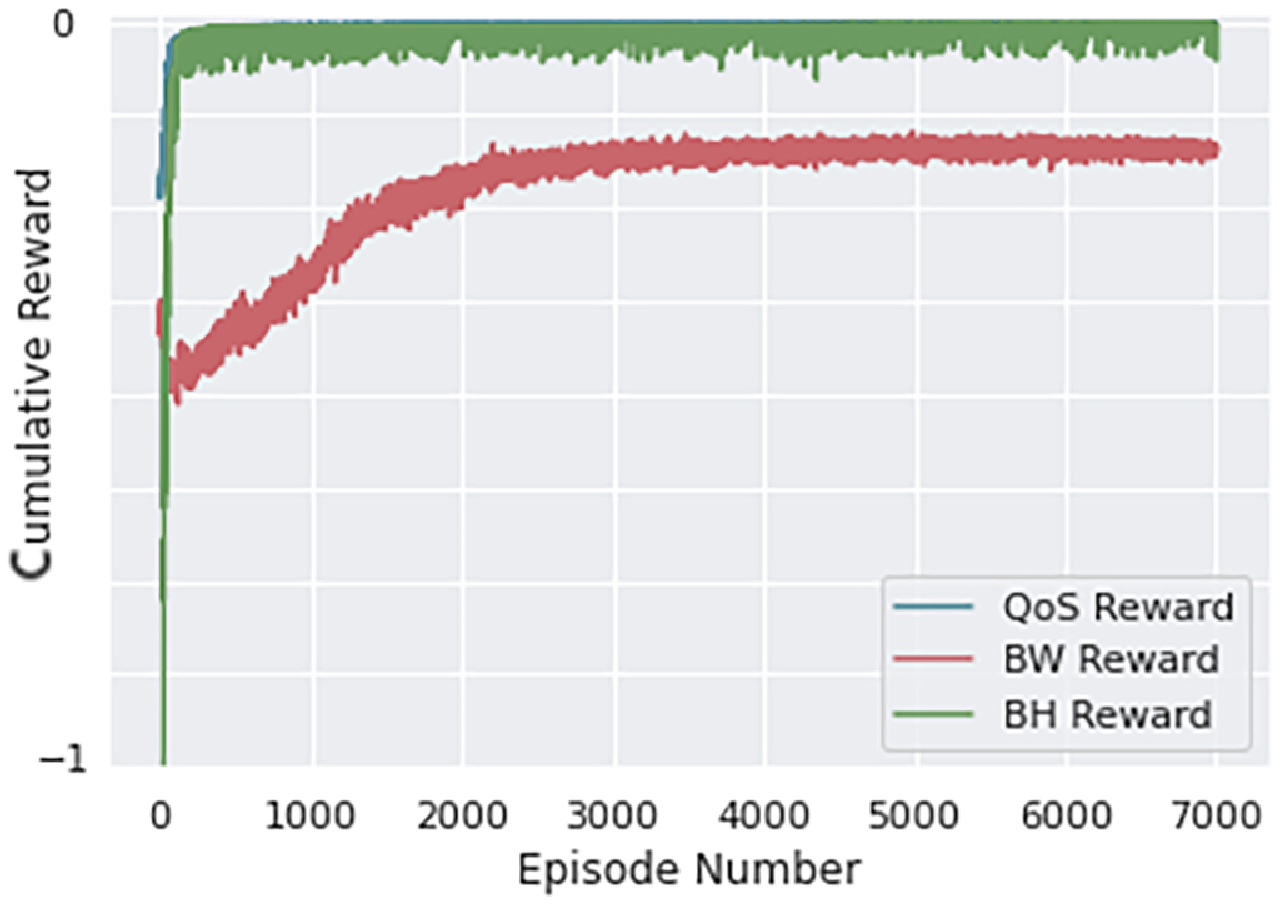}
		\captionof{figure}{Discounted cumulative reward for QoS \\metric, BW consumption and BH load obtained by \\ \textit{MO-A2C}.
			Rewards related to BW consumption and \\BH load are  multiplied by $10$ and $0.1$, respectively.	}
		\label{Fig_MOA2C_SD2}
	\end{minipage}%
	\begin{minipage}{.5\textwidth}
		\centering
		\hspace{-15 pt}\includegraphics[width=7.2 cm]{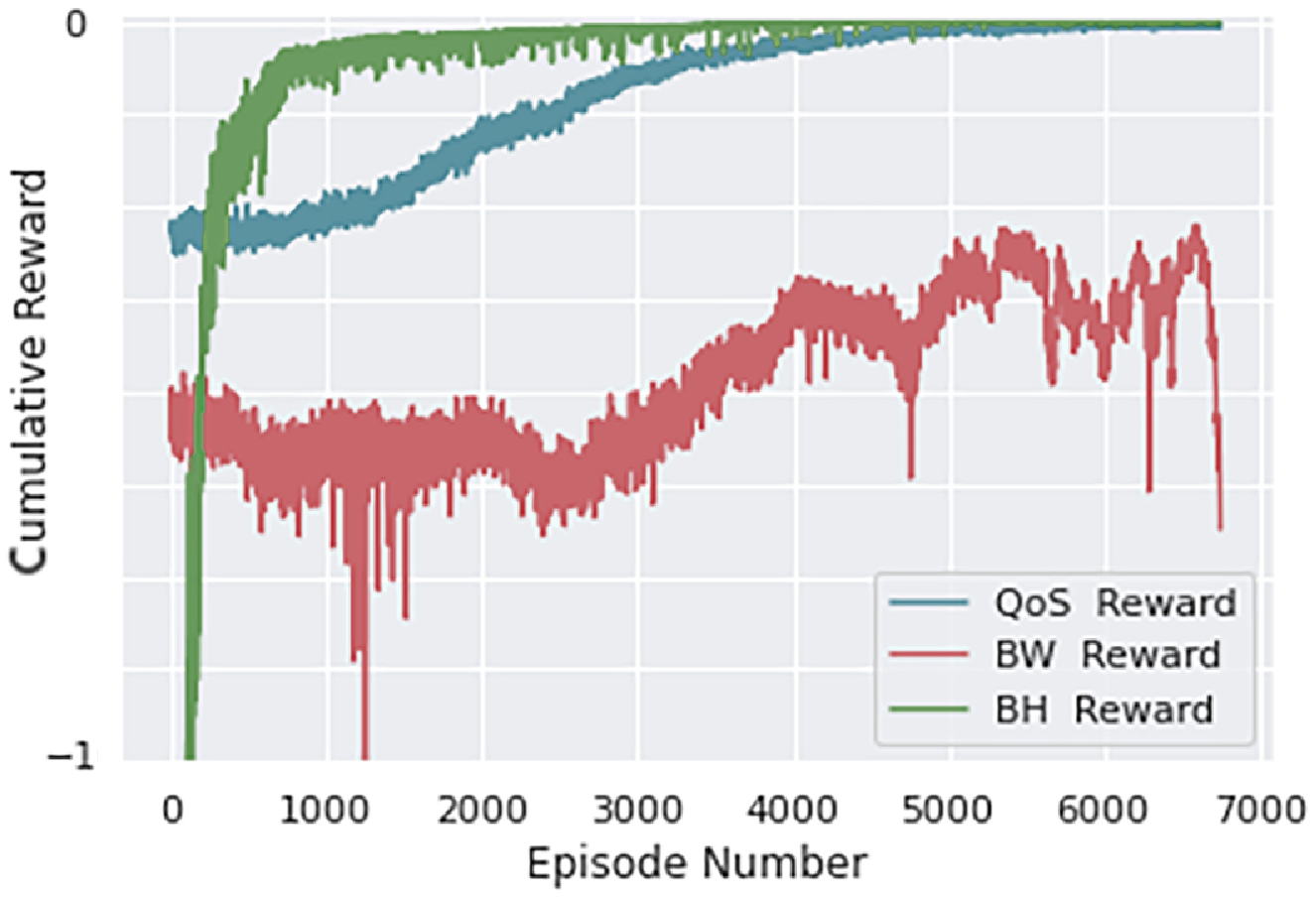}
		\captionof{figure}{Discounted cumulative reward for QoS \\metric, BW consumption and BH load obtained by \\ \textit{SO-A2C}.
			To constitute a scalarized reward, we use \\ the scales $[\lambda_{\rm QoS}, \:\lambda_{\rm BW}, \:\lambda_{\rm BH}] = [1,\:10,\:0.1]$.}
		\label{Fig_SOA2C_SD2}
	\end{minipage}
\end{figure}

We also consider another test scenario for scale-invariance evaluation of \textit{MO-A2C}.
For \textit{SO-A2C}, the scales $[\lambda_{\rm QoS}, \:\lambda_{\rm BW},\: \lambda_{\rm BH}] = [1,\:10,\:0.1]$ are used
and for \textit{MO-A2C}, we multiply the rewards related to BW consumption and BH load by $10$ and $0.1$, 
respectively, and keep the QoS reward unchanged. 
We then evaluate \textit{MO-A2C} over the re-scaled rewards.
The training performance of \textit{MO-A2C} and \textit{SO-A2C} 
are sketched in Figures \ref{Fig_MOA2C_SD2} and \ref{Fig_SOA2C_SD2},
in terms of the cumulative rewards for different episodes.
Comparison of Figures \ref{Fig_MOA2C_SD2} and \ref{Fig_SOA2C_SD2}, 
confirms that \textit{MO-A2C} is more robust to re-scaling factors than \textit{SO-A2C}
and again confirms that \textit{MO-A2C} can be considered as a scale-invariance approach.

\section{Conclusion}
In this paper, we devised a scale-independent multi-objective reinforcement learning approach
on the grounds of the advantage actor-critic (A2C) algorithm.
By making some assumptions, we then provided a convergence analysis based on which
a convergence-in-mean can be guaranteed.
We compared our algorithm for a multi-task problem against a single-objective A2C with a scalarized reward.
\textcolor{black}{The simulation results show the capability of the developed algorithm 
	from the sample-efficiency, optimality and scale-invariance perspectives}.

\section{Acknowledgement}
We thank Prof. Olav Tirkkonen %and Ashvin Srinivasan 
for the simulation matters.

\bibliography{iclr2022_conference}
\bibliographystyle{iclr2022_conference}

\appendix
\newpage
\textlarger{SUPPLEMENTARY MATERIAL}
\section{One Lemma Needed for Theorem \ref{Thm1}}
Before giving the proof of Theorem \ref{Thm1}, we need to present the following Lemma.

\begin{lem}\label{lemma1}
	Consider expected losses $\{{J}_{moa,j}(\cdot,\cdot)\}_1^r$ 
	and stochastic losses $\{\hat{J}_{moa,j}(\cdot,\cdot)\}_1^r$ complying with Assumptions 2 and 3, respectively,
	and $\boldsymbol{\alpha}_{\rm moa}$ being the solution of Eq. (\ref{EQ_MOAupdate}).
	Moreover, consider SGDes  (\ref{EQ_SGC_critic}) and  (\ref{EQ_SGC_actor}) characterized by 
	iteration number $i$ and 
	MO-actor learning rate $\{\mu_i\}_{i\in\mathbb{N}}$ with
	$$ 
	\mu_i \leq \min\left\{ \frac{1}{L},\frac{1}{L \| \bf{B}\|} \mathbb{E}_{\boldsymbol{\theta}}\Big\{  \dfrac{1}{\boldsymbol{1}_r^{\top}\left( \nabla \boldsymbol{J}({\boldsymbol{\theta}})^\top \nabla \boldsymbol{J}({\boldsymbol{\theta}}) \right)^{-1} \boldsymbol{1}_r} \Big\} \right\},
	$$
	which generate sequences $\{ {\boldsymbol{\phi}}^i \}_{i\in\mathbb{N}}$ and  $\{ {\boldsymbol{\theta}}^i \}_{i\in\mathbb{N}}$, then we get:
	$$
	\mathbb{E}\left\{ \sum_{j=1}^r \alpha_{\rm moa,j}^i \big({J}_{moa,j}({\boldsymbol{\theta}}^{i+1},{\boldsymbol{\phi}})-{J}_{moa,j}({\boldsymbol{\theta}}^{i},{\boldsymbol{\phi}})\big) \right\} \leq 0, \qquad\quad \mbox{for}~{\boldsymbol{\phi}} \in \Phi.
	$$
\end{lem}
\begin{proof}
	According to the update rule Eq. (\ref{EQ_SGC_actor}), we have:
	$$
	{\boldsymbol{\theta}}^{i+1} = {\boldsymbol{\theta}}^i - \mu_i \nabla{\boldsymbol{\hat{J}}}^i \boldsymbol{\alpha}_{\rm moa}^i, 
	$$
	where %${\boldsymbol{\theta}}^i$ is the generated solution by $i$-th iteration of SGD  Eq. (\ref{EQ_SGC_actor} and  
	$\nabla {\boldsymbol{\hat{J}}}^i= [\nabla_{{\boldsymbol{\theta}}} {\hat{J}}_{\rm moa,1},\ldots,\nabla_{{\boldsymbol{\theta}}} {\hat{J}}_{\rm moa,r}]({\boldsymbol{\theta}}^i,{\boldsymbol{\phi}}^i)$.
	Based on Assumption 2, we obtain:
	\begin{align*}
		{J}_{moa,j}({\boldsymbol{\theta}}^{i+1},{\boldsymbol{\phi}}) - {J}_{moa,j}({\boldsymbol{\theta}}^i,{\boldsymbol{\phi}}) \leq 
		- \mu_i {\nabla {J_{\rm moa,j}}({\boldsymbol{\theta}}^i)}^\top \nabla \boldsymbol{\hat{J}}^i \boldsymbol{\alpha}_{\rm moa}^i +
		\frac{\mu_i L^2}{2} 
		{\boldsymbol{\alpha}_{\rm moa}^i}^\top {\nabla \boldsymbol{\hat{J}}^i}^\top \nabla \boldsymbol{\hat{J}}^i \boldsymbol{\alpha}_{\rm moa}^i.
	\end{align*}
	Considering $\sum_{j=1}^r \alpha_{\rm moa,j}^i = 1$, it reads:
	\begin{align}\label{EQ_lem2_aux}
		&\mathbb{E}\left\{ \sum_{j=1}^r \alpha_{\rm moa,j}^i \big({J}_{moa,j}({\boldsymbol{\theta}}^{i+1},{\boldsymbol{\phi}})-{J}_{moa,j}({\boldsymbol{\theta}}^{i},{\boldsymbol{\phi}})\big) \right\} \notag \\
		&\qquad \leq\: \mathbb{E}\left\{ \mathbb{E}\left\{ 	\frac{\mu_i L^2}{2} 
		{\boldsymbol{\alpha}_{\rm moa}^i}^\top {\nabla \boldsymbol{\hat{J}}^i}^\top \nabla \boldsymbol{\hat{J}}^i \boldsymbol{\alpha}_{\rm moa}^i -
		\mu_i {\boldsymbol{\alpha}_{\rm moa}^i}^\top{\nabla \boldsymbol{J}^i}^\top \nabla \boldsymbol{\hat{J}}^i \boldsymbol{\alpha}_{\rm moa}^i \:\Big|\: {\boldsymbol{\theta}}^i,{\boldsymbol{\phi}}^i \right\} \right\} \notag \\
		&\qquad \overset{a}\leq\: -\mu_i\Big(1-\frac{\mu_iL}{2}\Big)\mathbb{E}\left\{ 	{\boldsymbol{\alpha}_{\rm moa}^i}^\top {\nabla {\boldsymbol{J}}^i}^\top \nabla {\boldsymbol{J}}^i \boldsymbol{\alpha}_{\rm moa}^i  \right\} + \frac{\mu_i^2 L}{2}\| \bf{B}\|,
	\end{align}
	where (a) was obtained based on $\mathbb{E}\left\{ {\boldsymbol{\alpha}_{\rm moa}^i}^\top{\nabla \boldsymbol{J}^i}^\top \nabla \boldsymbol{\hat{J}}^i \boldsymbol{\alpha}_{\rm moa}^i \:\big|\: {\boldsymbol{\theta}}^i,{\boldsymbol{\phi}}^i \right\} = {\boldsymbol{\alpha}_{\rm moa}^i}^\top{\nabla \boldsymbol{J}^i}^\top \nabla {\boldsymbol{J}}^i \boldsymbol{\alpha}_{\rm moa}^i$, 
	Assumption 3 and ${\boldsymbol{\alpha}_{\rm moa}^i}^\top \bf{B} \boldsymbol{\alpha}_{\rm moa}^i \leq \|\bf{B}\|\: \|\boldsymbol{\alpha}_{\rm moa}^i\|^2 \leq \|\bf{B}\|$.
	From Eq. (\ref{EQ_MOAupdate}), for all $\alpha_{\rm moa,j}^i \geq 0$, it reads:
	\begin{align*}
		\boldsymbol{\alpha}_{\rm moa}^i = \frac{ \left( \nabla {{\boldsymbol{J}}^i}^\top \nabla {\boldsymbol{J}}^i \right)^{-1} \boldsymbol{1}_r }
		{\boldsymbol{1}_r^{\top}\left( \nabla {{\boldsymbol{J}}^i}^\top \nabla {\boldsymbol{J}}^i \right)^{-1} \boldsymbol{1}_r}.
	\end{align*}
	By substituting this into Eq. (\ref{EQ_lem2_aux}), we get:
	\begingroup\makeatletter\def\f@size{9.5}\check@mathfonts
	\begin{align*}
		\!\!\!\mathbb{E}\bigg\{ \sum_{j=1}^r \alpha_{\rm moa,j}^i \big({J}_{moa,j}({\boldsymbol{\theta}}^{i+1},{\boldsymbol{\phi}})-{J}_{moa,j}({\boldsymbol{\theta}}^{i},{\boldsymbol{\phi}})\big) \bigg\} 
		\leq& -\Big(\mu_i-\frac{\mu_i^2 L}{2}\Big)\mathbb{E}\left\{ 	\Big(\boldsymbol{1}_r^{\top}\left( \nabla {{\boldsymbol{J}}^i}^\top \nabla {\boldsymbol{J}}^i \right)^{-1} \boldsymbol{1}_r\Big)^{-1}  \right\} \\
		&+ \frac{\mu_i^2 L}{2}\| \bf{B}\| \\
		\overset{a}\leq& -\frac{\mu_i}{2}\mathbb{E}\left\{ 	\Big(\boldsymbol{1}_r^{\top}\left( \nabla {{\boldsymbol{J}}^i}^\top \nabla {\boldsymbol{J}}^i \right)^{-1} \boldsymbol{1}_r\Big)^{-1}  \right\} \\
		&+ \frac{\mu_i^2L}{2}\| \bf{B}\| ~\leq~ 0,
	\end{align*}
	\endgroup
	where we used $\mu_i L \leq 1$ for (a).
	Considering that the denominator of RHS of the recent equation is positive due to the positive-definiteness of $\left( \nabla {{\boldsymbol{J}}^i}^\top \nabla {\boldsymbol{J}}^i \right)^{-1}$, the statement follows for $\mu_i < \frac{2}{L}$.
\end{proof}
\textcolor{black}{Lemma \ref{lemma1} guarantees that the expected value of scalarized loss $\mathbb{E} \left\{\sum_{j=1}^r \alpha_{\rm moa,j} J_{\rm moa,j}({\boldsymbol{\theta}}^i,{\boldsymbol{\phi}})\right\}$ monotonically reduces as the iteration increases, for ${\boldsymbol{\phi}} \in \Phi$.}
\begin{corollary}\label{Cor2} 
	Consider the framework of Lemma \ref{lemma1},
	then we get:
	$$
	\mathbb{E}\left\{ {\boldsymbol{\alpha}_{\rm moa}^i}^\top {\nabla {\boldsymbol{J}}^i}^\top \nabla {\boldsymbol{J}}^i \boldsymbol{\alpha}_{\rm moa}^i \right\} 
	\leq  \frac{2}{\mu_i} \mathbb{E}\left\{ \sum_{j=1}^r \alpha_j^i \big({J}_{\rm moa,j}({\boldsymbol{\theta}}^i,{\boldsymbol{\phi}})-{J}_{\rm moa,j}({\boldsymbol{\theta}}^{i+1},{\boldsymbol{\phi}})\big) \right\} + \mu_i L \| \bf{B}\|,
	$$
	for ${\boldsymbol{\phi}} \in \Phi$.
\end{corollary}
\begin{proof}
	According to Eq. (\ref{EQ_lem2_aux}) and $\mu_i L  \leq 1$, the statement follows.
\end{proof}

\section{Proof of Theorem \ref{Thm1}}
\begin{proof}
	Based on SGD update  (\ref{EQ_SGC_actor}), we obtain:	
	\begingroup\makeatletter\def\f@size{9.75}\check@mathfonts
	\begin{align*}
		\mathbb{E}\| {\boldsymbol{\theta}}^{i+1} - {\boldsymbol{\theta}}^* \|^2 =& \mathbb{E}\|{\boldsymbol{\theta}}^{i} - {\boldsymbol{\theta}}^* - \mu_i \nabla{\boldsymbol{\hat{J}}}^i \boldsymbol{\alpha}_{\rm moa}^i \|^2 \\
		\leq&~\mathbb{E}\| {\boldsymbol{\theta}}^{i} - {\boldsymbol{\theta}}^* \|^2 -2\mu_i  \mathbb{E}\bigg\{\mathbb{E}\Big\{ \sum_{j=1}^r \alpha_{\rm moa,j}^i \nabla \hat{J}_{\rm moa,j} ({\boldsymbol{\theta}}^i,{\boldsymbol{\phi}}^i)^\top\big( {\boldsymbol{\theta}}^{i}-{\boldsymbol{\theta}}^*\big) \: \Big|\: {\boldsymbol{\theta}}^i,{\boldsymbol{\phi}}^i \Big\} \bigg\} \\
		&+\mathbb{E}\left\{ \mu_i^2  {\boldsymbol{\alpha}_{\rm moa}^i}^\top {\nabla {\boldsymbol{\hat{J}}}^i}^\top \nabla {\boldsymbol{\hat{J}}}^i \boldsymbol{\alpha}_{\rm moa}^i \right\} \\
		\overset{a}\leq&~  (1-\gamma\mu_i)\:\mathbb{E}\| {\boldsymbol{\theta}}^{i} - {\boldsymbol{\theta}}^* \|^2 +2\mu_i \mathbb{E}\bigg\{\sum_{j=1}^r \alpha_{\rm moa,j}^i\big( J_{\rm moa,j}({\boldsymbol{\theta}}^*,{\boldsymbol{\phi}}^i) - J_{\rm moa,j}({\boldsymbol{\theta}}^i,{\boldsymbol{\phi}}^i)\big)\bigg\}\\
		&+\mu_i^2 \mathbb{E}\left\{\mathbb{E}\Big\{  {\boldsymbol{\alpha}_{\rm moa}^i}^\top {\nabla {\boldsymbol{\hat{J}}}^i}^\top \nabla {\boldsymbol{\hat{J}}}^i \boldsymbol{\alpha}_{\rm moa}^i \:\Big|\: {\boldsymbol{\theta}}^i,{\boldsymbol{\phi}}^i \Big\}  \right\}\\
		~\overset{b}\leq&~ (1-\gamma \mu_i)\: \mathbb{E}\| {\boldsymbol{\theta}}^i - {\boldsymbol{\theta}}^* \|^2 - 2\mu_i \mathbb{E}\bigg\{\sum_{j=1}^r \alpha_{\rm moa,j}^i \big({J}_{moa,j}({\boldsymbol{\theta}}^{i+1},{\boldsymbol{\phi}}^i)-{J}_{moa,j}({\boldsymbol{\theta}}^*,{\boldsymbol{\phi}}^i)\big)\bigg\} \\
		&+2\mu_i^2\| \bf{B}\|\\
		~\overset{c}\leq&~(1-\gamma \mu_i)\:\mathbb{E}\| {\boldsymbol{\theta}}^i - {\boldsymbol{\theta}}^* \|^2 + 2\mu_i^2 \| \bf{B}\|,
	\end{align*}
	\endgroup
	where (a) was obtained based on Assumption 2 and (b) according to Assumption 3 and Corollary \ref{Cor2}.
	For (c), we used Lemma \ref{lemma1} and $\alpha_{\rm moa,j}^i \geq 0$. 
\end{proof}
\end{document}